\documentclass{article}

\usepackage{microtype}
\usepackage{graphicx}
\usepackage{subfigure}
\usepackage{booktabs} % for professional tables
\usepackage{url}
\usepackage{amsmath, amsfonts, amsthm, amssymb}
\usepackage{bbm}
\usepackage{graphicx,subfigure}

\usepackage{wrapfig}
\usepackage{hyperref}

\usepackage[accepted]{icml2021}
\usepackage[,numbers]{natbib}

\begin{document}

\twocolumn[
\icmltitle{Doing Great at Estimating CATE? On the Neglected Assumptions in Benchmark Comparisons of Treatment Effect Estimators}%On the Neglected Assumptions in Benchmark Comparisons of Heterogeneous Treatment Effect Estimators}

%On the neglected assumptions underlying ML benchmark data sets for heterogeneous treatment effect estimation

% It is OKAY to include author information, even for blind
% submissions: the style file will automatically remove it for you
% unless you've provided the [accepted] option to the icml2021
% package.

% List of affiliations: The first argument should be a (short)
% identifier you will use later to specify author affiliations
% Academic affiliations should list Department, University, City, Region, Country
% Industry affiliations should list Company, City, Region, Country

% You can specify symbols, otherwise they are numbered in order.
% Ideally, you should not use this facility. Affiliations will be numbered
% in order of appearance and this is the preferred way.

\begin{icmlauthorlist}
\icmlauthor{Alicia Curth}{to}
\icmlauthor{Mihaela van der Schaar}{to,goo,ed}

\end{icmlauthorlist}

\icmlaffiliation{to}{University of Cambridge}
\icmlaffiliation{goo}{UCLA}
\icmlaffiliation{ed}{The Alan Turing Institute}

\icmlcorrespondingauthor{Alicia Curth}{amc253@cam.ac.uk}

% You may provide any keywords that you
% find helpful for describing your paper; these are used to populate
% the "keywords" metadata in the PDF but will not be shown in the document
\icmlkeywords{Machine Learning, ICML}

\vskip 0.3in
]

% this must go after the closing bracket ] following \twocolumn[ ...

% This command actually creates the footnote in the first column
% listing the affiliations and the copyright notice.
% The command takes one argument, which is text to display at the start of the footnote.
% The \icmlEqualContribution command is standard text for equal contribution.
% Remove it (just {}) if you do not need this facility.

\printAffiliationsAndNotice{}  % leave blank if no need to mention equal contribution
%\printAffiliationsAndNotice{\icmlEqualContribution} % otherwise use the standard text.

\begin{abstract} The machine learning toolbox for estimation of heterogeneous treatment effects from observational data is expanding rapidly, yet many of its algorithms have been evaluated only on a very limited set of semi-synthetic benchmark datasets. In this paper, we show that even in arguably the simplest setting -- estimation under ignorability assumptions -- the results of such empirical evaluations can be misleading if (i) the \textit{assumptions underlying the data-generating mechanisms} in benchmark datasets and (ii) \textit{their interplay with baseline algorithms} are inadequately discussed. We consider two popular machine learning benchmark datasets for evaluation of heterogeneous treatment effect estimators -- the IHDP and ACIC2016 datasets -- in detail. We identify problems with their current use and highlight that the inherent characteristics of the benchmark datasets favor some algorithms over others -- a fact that is rarely acknowledged but of immense relevance for interpretation of empirical results. We close by discussing implications and possible next steps.
\end{abstract}

\section{Introduction}
Successful estimation of treatment effects (TEs) from observational data is contingent on a range of assumptions on underlying data-generating processes (DGPs). Arguably the most essential class of assumptions in causal inference is concerned with \textit{identification} of effects. Such assumptions, e.g. the strong ignorability conditions \cite{rosenbaum1983central} or the backdoor criterion \cite{pearl1995causal}, render a TE identifiable -- yet they usually have no \textit{testable} implications, making their verification a task requiring domain expertise \cite{pearl2009causality}. Another class of assumptions, which is less often the source of controversy, is \textit{statistical} in nature and mainly linked to \textit{estimation performance}; e.g. assumptions on overlap, error terms, and -- in particular -- the form and complexity of underlying regression functions in the DGP determine the (expected) relative performance of different algorithms on a dataset. While the former class thus fundamentally ensures that \textit{any} estimate can be interpreted as causal, the latter class of assumptions is crucial in practice when it comes to \textit{choosing} an algorithm from the ever-expanding machine learning (ML) toolbox for heterogeneous treatment effect (HTE) estimation. 

When new algorithms are proposed within the ML community, it is common to evaluate their performance against existing \textit{baseline algorithms} on \textit{benchmark datasets}, which, in other areas of ML, often consist of real data e.g. ImageNet \cite{deng2009imagenet}. When trying to evaluate estimators of HTEs on real data, the fundamental problem is that -- even if identifying assumptions hold -- ground truth individual TEs are never observed \cite{holland1986statistics}. Thus, to provide proof-of-concept, papers proposing new (H)TE estimators have largely relied on synthetic or semi-synthetic datasets to showcase their properties. Possibly to discourage authors to invent DGPs that put their own algorithms in the best light, a few of these (semi-synthetic) datasets have emerged as benchmarks which have been used in a wide range of ML papers in the last years. The most prominent example of this is \cite{hill2011bayesian}'s IHDP dataset; since its first use in \cite{johansson2016learning, shalit2017estimating}, it has become the standard benchmark dataset in the ML HTE literature (used in e.g. \cite{alaa2017bayesian, alaa2018limits, hassanpour2019counterfactual, hassanpour2020learning, yoon2018ganite, assaad2020counterfactual, curth2020, zhang2020learning}). Unfortunately, this reliance on standard benchmark datasets and common baseline algorithms has seemingly removed the pressure on authors to (i) examine assumptions underlying such DGPs and (ii) argue why a specific benchmark comparison is fair and insightful in any given context. 

\textbf{Motivating example.} A popular baseline algorithm for HTE estimation from the statistics community is \cite{wager2018estimation, athey2019generalized}'s Causal Forest (CF), which -- unlike most ML estimators -- targets the TE directly, comes with a set of theoretical guarantees and has already been applied in real empirical studies \cite{athey2019estimating, davis2017using}. Nonetheless, \cite{shalit2017estimating} (and many extensions, e.g.  \cite{assaad2020counterfactual, johansson2018learning, yao2018representation}) show that the neural network (NN) based TARNet and its many variants outperform CF by lengths on the IHDP dataset. We asked ourselves a simple question: \textit{Why?} Is it because TARNet is a uniformly better estimator? Or is it because TARNet is a NN while CF is a random forest (RF)? Is it maybe because TARNet models the potential outcomes, while CF models the TE directly? Or is it something else inherent to DGP or implementation? Such questions are by no means unique to this specific example and arise in many more ML HTE papers -- in fact, missing insights into the sources of performance differences is a problem of the ML community at large \cite{lipton2018troubling}. Therefore, these questions only illustrate that it is crucial to build better understanding of how to interpret empirical findings in light of the assumptions underlying both benchmark datasets and baseline algorithms.

\textbf{Outlook.}  In this paper, we argue that the recent ML literature on HTE estimation has largely neglected how assumptions, inherent to the semi-synthetic DGPs used for testing, influence benchmark comparisons of newly proposed and baseline algorithms. To substantiate this, we present two case studies using popular benchmark datasets -- the IHDP dataset \cite{hill2011bayesian} and the ACIC2016 simulations \cite{dorie2019automated} -- in which we compare the empirical performance of a number of RF- and NN-based HTE estimators as illustrative examples. By doing so, we aim to understand to what extent some observed performance differences are simply a product of the interplay between (i) the assumptions underlying the semi-synthetic DGPs of the datasets and (ii) the assumptions underlying the compared algorithms, giving some algorithms an \textit{expected} advantage that should be disclosed if the goal is fair comparison. %This is particularly crucial in the TE estimation context because current benchmarks are semi-synthetic due to absence of ground truth in real datasets. 
In passing, we also highlight other problems with the current execution of benchmark comparisons. % with how these benchmark datasets are currently used: because these datasets are comprised of multiple runs using \textit{different DGPs}, averaging over all runs masks interesting performance differences. 
Ultimately, we aim to build greater understanding of what such (semi-synthetic) benchmark comparisons \textit{can and cannot establish}, and therefore close by discussing practical implications for the ML HTE literature.

\section{Problem Setup, Assumptions and Their Interplay with Learning Algorithms}
In this paper, we operate under the standard setup in the potential outcomes (PO) framework \cite{rubin2005causal}. That is, we assume that any individual, associated with (pre-treatment) covariates $X \in \mathcal{X}$, has two potential outcomes $Y(0)$ and $Y(1)$ of which only $Y=Y(W)$, the outcome associated with the administered (binary) treatment $W \in \{0, 1\}$, is observed. We are interested in $\tau(x)$, the conditional average treatment effect (CATE), which is the expected difference between an individual's POs (conditional on covariates), i.e.
\begin{equation}\label{eq:catedef}
    \tau(x) = \mathbb{E}[Y(1) - Y(0)|X=x] = \mu_1(x) - \mu_0(x)
\end{equation}
where $\mu_w(x)=\mathbb{E}[Y(w)|X=x]$ is the expected PO. 

\textbf{Identifying assumptions.} We aim to reason about the expected properties of a number of ML-based (or nonparametric) estimators of CATE, all of which rely on the effect being identified. As our main interest here lies \textit{not} in identification but in a different class of assumptions (see below), we therefore rely on the strong ignorability conditions \cite{rosenbaum1983central} for convenience (with the understanding that they may be limiting in practical applications).

\subsection{Key assumptions on the POs} Instead, our core interest lies in how assumptions on the form of and relationship between the two PO regression surfaces affect the relative performance of different estimators. In DGPs used for testing CATE estimators, such assumptions manifest in how the $\mu_w(x)$ are modeled. In most general form, one can always let
\begin{equation}\label{eq:gendef}
    \mu_w(x) = \begin{cases}f_0(x) & \text{if} ~w=0\\
    f_1(x) &\text{if} ~w=1\end{cases}
\end{equation}
for $f_0, f_1$ some functions, which could be arbitrarily different. While possible in theory, having no relationship between the expected outcomes under different treatments seems highly unrealistic in practice. In medicine, for example, one often assumes that some biomarker information is \textit{prognostic} of outcome \textit{regardless} of treatment status, such that only a subset of markers is \textit{predictive} of effect heterogeneity \cite{ballman2015biomarker}. A common approach to simulating response surfaces is therefore to simply use additive effects,
\begin{equation}
    \mu_w(x) = f_0(x) + w f_\tau(x)
\end{equation}
that is, to assume that the treated regression surface $f_1(x)$ in (\ref{eq:gendef}) can be \textit{additively decomposed} into a component shared with the control group ($f_0(x)$) and $f_\tau(x)$, an offset function determining treatment effect and heterogeneity, which could be simpler (e.g. smoother or sparser) than $f_1(x)$ itself.%\footnote{Note that, by the definition of CATE in (\ref{eq:catedef}), it is clearly always possible to write $\mu_1(x)=\mu_0(x)+w\tau(x)=\mu_0(x)+w(\mu_1(x)-\mu_0(x))$ even if $f_0$ and $f_1$ are completely unrelated, however, we would not qualify this as an additive decomposition as in this case $f_\tau(x)$ would be more complex than $f_1(x)$ itself.} 

Instead of additive transformations, it would also be possible that the PO regression surfaces have more general relationships, e.g. $f_1(x)=g(f_0(x))$ with $g(\cdot)$ some transformation function -- in the IHDP dataset, for example, $g(\cdot)$ is a logarithmic transformation -- however, this specification seems less popular than additive parametrizations used in most DGPs in related work. 

\subsection{Learning Algorithms and the interplay with DGPs} A plethora of ML-based CATE estimators have been proposed in recent years. Here, we distinguish them along two key axes most relevant to our problem: (i) the underlying \textit{ML method} and (ii) the \textit{estimation strategy}. The former is straightforward and refers simply to the ML method used to implement an algorithm, e.g. a NN or a RF. The latter is crucial in the CATE context but has received relatively little explicit attention in the ML literature. As in \cite{curth2020}, we distinguish between two types of estimation strategy: \textit{indirect} estimators that target the POs, i.e. first obtain estimates $\hat{\mu}_w(x)$ and then simply set $\hat{\tau}(x)=\hat{\mu}_1(x)-\hat{\mu}_0(x)$, and estimators that target CATE \textit{directly}, e.g. by using pseudo-outcome regression or other two-step procedures \cite{kunzel2019metalearners, nie2017quasi, kennedy2020optimal, curth2020}. 

From a theoretical viewpoint, we expect that the type of underlying DGP will (at least partially) determine which combination of ML method and estimation strategy will be most successful on any benchmark dataset. To see this, let $\textstyle{\epsilon_{sq}(\hat{f}(X)) = \mathbbm{E}[(\hat{f}(X)\!-\!f(X))^2]}$ denote the expected MSE for an estimate $\textstyle \hat{f}(x)$ of a function $\textstyle f(x)$ and consider the behaviour of $\textstyle{\epsilon_{sq}(\hat{\tau}(X)}$ for different strategies. For indirect estimators, we have that $\textstyle{\epsilon_{sq}(\hat{\tau}(X)) \leq 2(\epsilon_{sq}(\hat{f}_1(X))+\epsilon_{sq}(\hat{f}_0(X)))}$, so that the error rate on the more complex of the regression surfaces will determine performance \cite{alaa2018limits, curth2020}. Some direct estimators, on the other hand, can instead reach the same performance as a supervised learning algorithm with (hypothetical) target $Y(1)-Y(0)$, so that they can attain the error rate associated with $f_\tau(x)$ (see \cite{curth2020,kennedy2020optimal,kunzel2019metalearners}). Whenever $f_\tau(x)$ is easier to estimate %\footnote{Depending on assumptions on the underlying functions, `simpler' could mean e.g. \textit{sparser} or \textit{smoother} in this context (as in e.g. \cite{kennedy2020optimal, curth2020}).} 
than the more complex of $f_1(x)$ and $f_0(x)$ (e.g. due to being sparser or smoother \cite{curth2020, kennedy2020optimal}), direct learners thus have a clear theoretical advantage -- which diminishes as $f_1(x)$ and $f_0(x)$ become less similar (making $f_\tau(x)$ increasingly complex). Further, the properties of the underlying ML method used to implement any estimation strategy will determine how well different types of functions $f_w(x)$ and/or $f_\tau(x)$ can be fit using a finite sample of observed data from a specific DGP. 

\section{Empirical investigation}
In this section we present two case studies comparing the empirical performance of estimators relying on (i) different ML-methods (NNs and RFs\footnote{In principle, many other base-methods could have been considered for this study; other popular CATE estimators rely on BART \cite{hill2011bayesian, hahn2017bayesian}, Gaussian Processes \cite{alaa2017bayesian, alaa2018limits} and GANs \cite{yoon2018ganite}. Here, we focus on RFs due to their popularity in applied research and NNs due to their popularity in the recent ML HTE literature.}) and (ii) different estimation strategies (direct and indirect). We begin with our motivating example and examine the sources of performance differences between the RF-based, direct estimator Causal forest (CF) \cite{wager2018estimation, athey2019generalized} and the NN-based, indirect estimator TARNet\footnote{Here, we focus on \cite{shalit2017estimating}`s TARNet instead of CFRNet (which adds a balancing term to handle selection on observables) because the latter adds an \textit{additional modeling dimension} on top of the two dimensions whose effect we wish to isolate in this paper.} \cite{shalit2017estimating} on the IHDP dataset. Second, we consider relative performance of the same algorithms on a subset of the ACIC2016 simulations.

\textbf{Models and implementation. } We consider two (model-agnostic) indirect estimation strategies which are commonly referred to as T- and S-learner \cite{kunzel2019metalearners}; the former fits \textbf{t}wo separate regression surfaces for each treatment arm, while the latter fits a \textbf{s}ingle model in which $W$ is included as a standard covariate. We consider standard RF-based implementations (\textbf{TRF} and \textbf{SRF}) and a NN-based T-learner (\textbf{TNet}). Instead of a standard NN-based S-learner we use \textbf{TARNet} (as it can be seen as a hybrid of S- and T-learner \cite{curth2020}). As direct estimators, we consider \cite{athey2019generalized}'s \textbf{CF}, which relies on a two-stage procedure solving a local moment equation inspired by the Robinson transformation \cite{robinson1988root}, and, as a NN-based variant, we use \cite{nie2017quasi}'s R-learner (\textbf{RNet}), which relies on the same principle. For all forest-based methods we use the R-package \url{grf} \cite{grf} and for all NN-based methods we use the python-implementations \url{catenets} \cite{curth2020}. We use all models off-the-shelf; for all RFs this entails using 2000 trees, and all NNs have hyperparameter settings similar to those used in \cite{shalit2017estimating} for the IHDP experiments. For further details, refer to the the Appendix. 

\subsection{Case study 1: IHDP dataset}
The IHDP benchmark dataset uses a  semi-synthetic DGP on top of the covariates ($n=747, d=25$) and treatments of the Infant Health and Development Program, a randomized experiment targeting an intervention at premature infants with low birth weight. \cite{hill2011bayesian} introduced selection bias and imbalance ($n_0=608, n_1=139$) by excluding a non-random proportion of treated individuals (those with nonwhite mothers), leading to incomplete overlap for the control group. The popular semi-synthetic DGP (setup `B' in \cite{hill2011bayesian}) uses $\mu_0(x)=\exp((x+A)\beta)$ and $\mu_1(x)=x\beta - \omega$; here $\beta$ is a coefficient vector with entries sampled from $(0, 0.1, 0.2, 0.3, 0.4)$ with probabilities $(0.6, 0.1, 0.1, 0.1, 0.1)$, $A$ is a fixed offset matrix, and $\omega$ is set uniquely in every simulation run, ensuring that the average treatment effect on the treated (ATT),  which was the main estimand of interest in \cite{hill2011bayesian}, is equal to 4. Note that this DGP was \textit{not} created to mimic a specific realistic response surface, rather it was one of multiple DGPs used in \cite{hill2011bayesian} to compare different estimators. Below, we use \cite{shalit2017estimating}'s IHDP-100 dataset (100 draws of the DGP) and report performance on the pre-determined 10\% hold-out sample. We repeat each run 5 times with different seeds for all models.
 
\subsubsection{Empirical results}
\begin{figure*}[!t]
	\centering
	\subfigure[Original IHDP]{\label{fig:ihdp_main}\includegraphics[width=0.9\columnwidth]{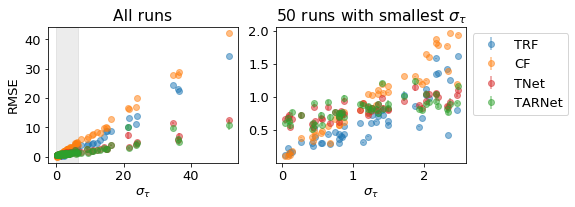}}
    \subfigure[Modified IHDP (Additive TE)]{\label{fig:ihdp_mod}\includegraphics[width=0.9\columnwidth]{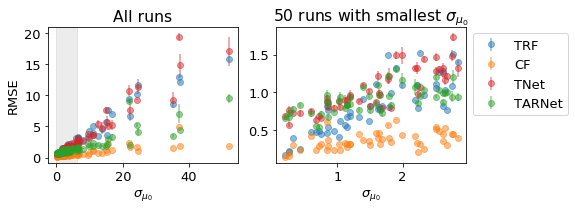}}
    \vspace{-4.5mm}
	\caption{Out-of-sample RMSE of CATE estimation across 100 IHDP draws (original and modified setting)  for TNet, TARNet, TRF and CF. Averaged across 5 runs, bar indicates one standard error. Shaded area in left plots indicates area which are zoomed on in right plots.}\label{fig:ihdp_all}      
\end{figure*}

\begin{wrapfigure}[8]{r}{0.47\columnwidth}
    \vskip -0.3in
    \centering
    \includegraphics[width=0.45\columnwidth]{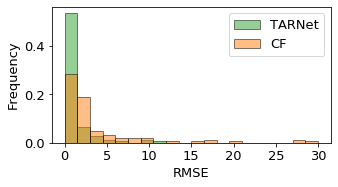}
      \vskip -0.05in
    \caption{Histogram of test RMSE of CATE estimation for TARNet and CF on IHDP}
    \label{fig:hist}
    \vskip -0.15in
\end{wrapfigure}\textbullet \textbf{Finding (i): Reporting simple averages of RMSE across simulation runs appears inappropriate.} We begin with a general observation we consider crucial for anyone using the IHDP dataset for benchmark comparisons. In Fig. \ref{fig:hist} we plot a histogram of the average RMSE across the 100 realizations of the simulation, in which it is obvious that the right tail of scores very long and heavy. Averaging over such scores gives extremely high weight to the few runs with very high RMSE, which, as we show below, are the runs with large variation in CATE. This is because \cite{hill2011bayesian} fixed the magnitude of the ATT across runs, but CATE is allowed to vary freely, leading to $\sigma_{\tau}=\sqrt{Var_{\text{test}}(\tau(x))}$ (capturing the spread of CATE within the test-set of a run) differing by orders of magnitude, ranging between $0.04$ and $51.51$, across the 100 simulations. The common practice to simply report an average RMSE score across all runs gives algorithms that perform well mainly in the tails a clear advantage; thus it might be more appropriate to consider alternative metrics (e.g. normalized RMSE or paired sample test statistics).

\textbullet \textbf{Finding (ii): Indirect strategies perform best on this dataset and NNs outperform RFs.} In the left panel of Fig. \ref{fig:ihdp_main}, we plot RMSE by $\sigma_{\tau}$ for TARNet, TNet, TRF and CF on IHDP-100. We omit RNet and SRF for readability, refer to the appendix for full results. We observe that the direct estimators consistently perform worse than their indirect alternatives. This is expected given the DGP and the theoretical arguments made in the previous section: $\mu_0(x)$ and $\mu_1(x)$ are not similar (on an additive scale) in the underlying DGP, making $\tau(x)$ a difficult function to estimate directly. Further, considering only the left panel of Fig. \ref{fig:ihdp_main}, it appears as if NN-based estimators have a clear advantage on this dataset and the underlying ML-method seems to matter much more than the CATE estimation strategy used. TARNet also consistently outperforms TNet; this is not surprising given that the linear predictor $X\beta$ is shared across both outcome surfaces, making for a perfect shared representation that TARNet can exploit. 

\textbullet \textbf{Finding (iii): Relative performance systematically differs across runs. } Note that simulation runs in which more covariates have (large) nonzero coefficients will have higher TE heterogeneity due to the exponential specification. In the left panel of Fig. \ref{fig:ihdp_main}, it is obvious that as the measured variation in CATE increases, the absolute discrepancy between methods becomes more extreme. Due to high differences in magnitude, this perspective masks more interesting differences that become most apparent when considering \textit{relative} differences using only the simulation runs with less extreme $\sigma_{\tau}$. In the right panel of Fig. \ref{fig:ihdp_main} we observe that the forest-based estimators perform \textit{better} than the NN-based versions for $\sigma_{\tau}$ small, and that the discrepancy between direct and indirect learners is less extreme.

\begin{figure}[!b]
\includegraphics[width=0.95\columnwidth]{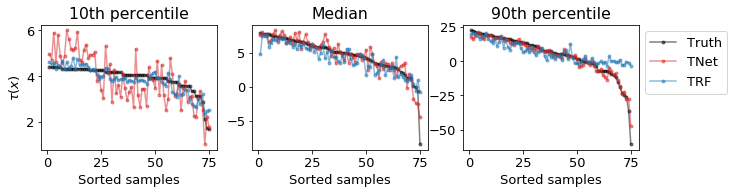}
\caption{True and predicted CATE for TNet and TRF, on 3 datasets with $\sigma_\tau$ at 10, 50 and 90th percentile across IHDP runs.}\label{fig:boundary}
\end{figure}

\textbullet \textbf{Finding (iv): Tree-based methods suffer in the tails due to the exponential.} The performance differences highlighted above let us speculate that the poor performance of the forest-based estimators (which can be thought of as adaptive nearest neighbor methods \cite{wager2018estimation}) may be partially caused by boundary bias on exponential specifications. We confirm this in Fig. \ref{fig:boundary} where we plot T-learner estimates of $\tau(x)$ for the simulation runs at the 10, 50 and 90th percentiles of $\sigma_\tau$. We observe that for large $\sigma_\tau$, major performance discrepancies indeed arise only for the small subset of individuals with the largest linear predictor, which is where the exponential is the steepest. 

\begin{figure*}[!t]
	\centering
    \includegraphics[width=\textwidth]{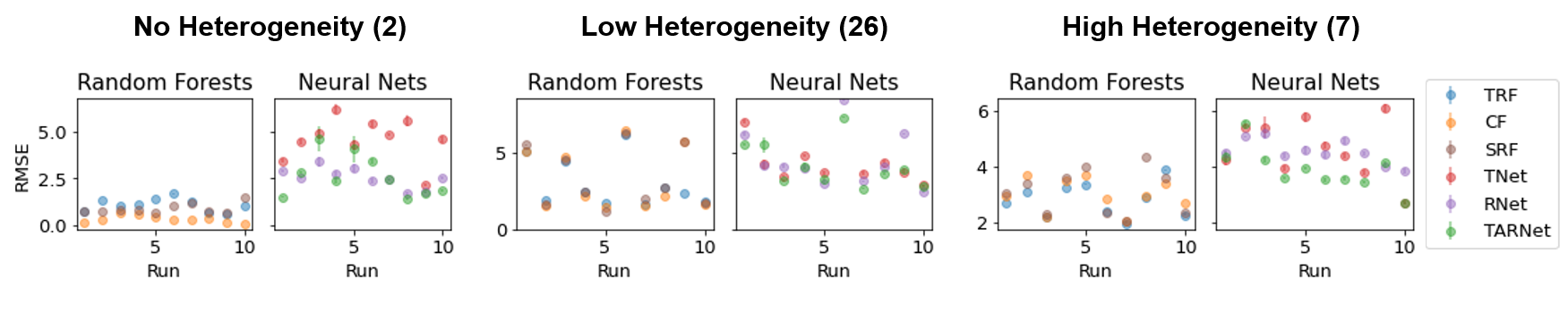}
    \vspace{-10mm}
	\caption{Out-of-sample RMSE of CATE estimation across 10 runs of a no, low, and high heterogeneity simulation setting. Averaged across 5 (10 runs for NNs) runs, bar indicates one standard error}\label{fig:acicall}     \vspace{-4.5mm}
\end{figure*}

\textbullet \textbf{Finding (v): Tweaking the DGP slightly by creating an additive TE leads to completely different results. } Finally, to further test finding (ii), i.e. whether the observed performance differences across strategies are indeed due to the CATE function being as difficult to estimate as the $\mu_w(x)$, we slightly alter the original IHDP simulation. We use $\mu^*_0(x)=\mu_0(x)$ and $\mu^*_1(x)=\mu_1(x)+\mu_0(x)$; i.e. the treatment effect is now additive and simple (linear). In Fig. \ref{fig:ihdp_mod} we report RMSE of estimating CATE by $\sigma_{\mu_0}$ (as we found the variance induced by the exponential specification in $\mu_0(x)$, and not $\sigma_{\tau}$, to drive variation across runs). We observe that in this setting, almost all conclusions on relative performance are indeed \textit{reversed} from what we observed in \ref{fig:ihdp_main}: CF performs best throughout, direct learners perform better and NNs no longer have a clear advantage over RFs (possibly because the boundary bias now appears on \textit{both} regression surfaces, which can difference out). 

\subsubsection{Conclusion case study 1} 
The empirical investigation above allows us to resolve our questions asked in the motivating example: The advantage of TARNet over CF on the IHDP dataset has multiple sources; the DGP underlying the IHDP dataset indeed favours both (i) NNs over RFs and (ii) indirect over direct estimators. Further, CF performs worse on runs with extreme $\sigma_{\tau}$, which effectively get much higher weight than the runs with very low $\sigma_{\tau}$ where it performs best. When tweaking the DGP slightly to create a simple and additive treatment effect, the relative performance reverses, highlighting that performance is indeed determined by the interplay between assumptions underlying a learning algorithm and the DGP.

%\textbf{Maybe shared representation?}
%Finding of interest: performance of TARNet/CFRNet versus causal forest. What drives poor performance of CF? Compare: Nets s.a. TARNet, R-learner with forests s.a. CF, S/T-learner. 
%Plot: (relative) performance against sd(CATE) or sd(y); to show both impact of one-vs-two-step learner and boundary bias for tree-based models

%Problems with IHDP data-set:
%- biased towards models that learn shared representations
%- biased towards models that somehow enforce sparsity
%- biased towards models which are good at estimating very complex treatment effects
%- biased towards models which can represent exponentials well (boundary bias...)
%- problematic to average across data

\subsection{Case study 2: ACIC2016}
The datasets used in the Atlantic Causal Inference Competition (ACIC) 2016 are based on real covariates  ($n=4802, d=58$) from the Collaborative Perinatal Project. The competition organizers created 77 simulation settings which varied in the functional form of the response surfaces, and the degrees of confounding, overlap and TE heterogeneity (see \cite{dorie2019automated} for more detail); also here the main goal was to estimate the ATT. These datasets were used in e.g. \cite{alaa2019validating, assaad2020counterfactual, jesson2020identifying, lu2020reconsidering} to evaluate CATE estimators. Once more, our main interest lies in how the PO specification in the DGP influences the observed relative performance of algorithms. Therefore, we consider only a subset of settings and fix all `experimental knobs' except for the degree of TE heterogeneity (which determines the similarity of the POs). Here, we focus on settings 2, 26 and 7, which have exponentials in their response surfaces and differ only in that they have no, low and high TE heterogeneity, respectively\footnote{We chose this triplet as it is the only one which has all three heterogeneity settings available.}. %We present settings with different response surfaces in the supplement. 
For each setting, we present out-of-sample results for the first 10 (out of 100) simulation runs provided by \cite{acic16}, where we use the first 4000 observations for training and the remaining 802 for testing. Because of their higher variability, we average all NN results across 10 replications instead of 5. 

\subsubsection{Empirical results} 
\textbullet \textbf{Finding (i): There is substantial variation in absolute performance across different runs of \textit{the same} setting.} We again make a general observation on distribution of RMSE scores. Similar to the IHDP dataset, there is substantial variation in absolute performance of algorithms across different runs \textit{of the same setting}. As can be seen in Fig. \ref{fig:acicall}, when considering the performance of the forest-based estimators, the differences in \textit{absolute} performance of \textit{the same} algorithm across different runs are often larger than the differences between different strategies (using the same underlying ML method) on the same run.\footnote{We conjecture that this is because \cite{dorie2019automated} randomly sample terms which enter the response surfaces in each run, making some runs randomly harder than others.}  Similar to the IHDP dataset, simple averaging across runs can therefore mask differences that are visible mainly on the run-level. 

\textbullet \textbf{Finding (ii): Relative performance of direct and indirect learners across heterogeneity settings varies as expected.} In Fig. \ref{fig:acicall}, we observe that there is a trend across the three heterogeneity settings: In absence of heterogeneity, direct learners have a clear advantage and S-learners outperform T-learners. For low heterogeneity, all methods show similar performance. For high heterogeneity, the observations are reversed and indirect learners generally perform best. This difference is much more consistent and clear for the tree-based estimators than for the NN-based estimators. Further, TARNet, as a hybrid between S- and T-learner strategy, seems to inherit both their advantages, and can even match the performance of RNet on the setting without HTE.

\begin{figure}[!t]
\vspace{-3mm}
\includegraphics[width=0.9\columnwidth]{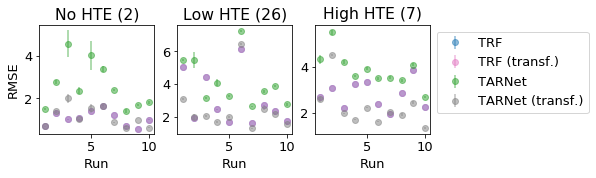}\vspace{-4mm}
\caption{Out-of-sample RMSE of CATE estimation for TRF and TARNet with and without pre-transformed data. Purple dots are a consequence of blue and pink dots overlapping perfectly.}\label{fig:acictrans}
\vspace{-5.5mm}
\end{figure}

\textbullet \textbf{Finding (iii): Underlying simulation favors tree-based methods.} Despite looking at a setting with exponential response surfaces and abundant data, we observe that in Fig. \ref{fig:acicall}, RFs outperform the NNs -- which stands in apparent contrast to the findings in Case study 1. Finally, we therefore investigated whether there is a reason for this, hidden in the DGP. We found that the simulated response surfaces are \textit{not} created using the raw data as input, instead the 27 count variables are dichotomized by \cite{dorie2019automated} before they are used in the DGP. This is the most natural pattern to represent for a RF, but we speculated that the NNs may struggle with this. To test this hypothesis, we feed the transformed data as used in the DGP by \cite{dorie2019automated} to both RFs and NNs. In Fig. \ref{fig:acictrans}, we observe that indeed the performance of the RFs does not change, while the NNs improve substantially. This highlights that also here, one ML method is a priori better equipped to model the response surfaces than another. 

% making it difficult
%Finding of interest: performance of NNet based models vs tree-based, CF vs T-learner RF

%Problems with ACIC2016:
%- Might be biased towards tree-based methods by the way the data is processed (have to check)
%- problematic to average across data?
%- so many settings that often blanket statements of performance across all settings are reported, which is not how it was designed
%- has overlap violations (was designed for estimation of ATT)

%(hyperparameter selection is done weirdly as well)

\subsubsection{Conclusion Case study 2}
We found that, in the considered simulations of ACIC2016, the relative performance of different learning strategies across heterogeneity settings is also in line with expectations. As the 77 settings provided by \cite{dorie2019automated} consist of only 2 settings without heterogeneity, 32 with low and 43 with high heterogeneity, we would expect that -- on average -- indirect learners will generally be favoured on this benchmark. Further, we found that some aspects of the underlying DGP may inherently favor tree-based methods.

\section{Conclusions and Implications} The main goal of this paper was to raise awareness of the limitations inherent to the current use of generic \textit{semi-synthetic} benchmark comparisons in the ML HTE literature. Our case studies highlighted that for semi-synthetic benchmark datasets in which (a component of) the DGP is known, some estimators have an \textit{expected} advantage over others, due to a better fit of underlying ML method and/or estimation strategy with the assumptions underlying the generation of the POs. Future research could consolidate these findings by investigating further datasets, other estimators and the effect of hyperparameters. 

Overall we do not consider our findings surprising -- they follow directly from theoretical reasoning about the properties of different algorithms. Nonetheless, such arguments are rarely taken into account (or are at least not explicitly discussed) when benchmark datasets and baseline algorithms are selected for use in related work. Therefore, we close by discussing implications, presented as `food-for-thought' in form of questions to the audience below.

\textbullet \textbf{\textit{What are sensible baseline algorithms for a proposed estimator?}} Because many DGPs will favour one class of algorithms over another, we consider it important for authors to provide \textit{principled} instead of `apples to oranges' baseline comparisons -- the latter may mask the sources of performance gain. That is, we consider some baseline comparisons more insightful than others. We would, for example, be interested whether a newly proposed \textit{learning strategy} outperforms existing learning strategies implemented using the \textit{same underlying ML method}, or conversely, whether a new ML method applied to the CATE estimation context outperforms commonly applied ML methods using the same (pre-existing) learning strategy (instead of direct comparisons of estimators that vary along \textit{multiple} dimensions).

\textbullet \textbf{\textit{What are insightful experimental knobs  for a proposed estimator?}} Because it is often possible to predict under which circumstances an algorithm will perform best, authors should be incentivized to demonstrate how the performance of their proposed algorithm changes as experimental knobs, capturing the relevant dimension, change from most to least favourable settings. Requiring the use of `standard' benchmark datasets can be detrimental in this context; especially when there are no relevant experimental knobs or settings (e.g. for evaluating direct estimators on the IHDP dataset). Conversely, when a standard benchmark dataset is used, a discussion of its inherent characteristics and their interplay with considered algorithms should be encouraged.

\textbullet \textbf{\textit{What does it mean for an algorithm to be `state-of-the-art'?}} As we highlighted in our experiments, the relative performance of different algorithms can change when they are compared under different DGPs/assumptions. This means that performance assessments based on a single dataset/DGP (e.g. IHDP) capture only one specific setting of many configurations of possible drivers of relative performance. We think that this warrants further discussion in the community reconsidering the meaning of the label `state-of-the-art'.

\textbullet \textbf{\textit{Can we create better benchmark datasets?}} Here, we considered only two (highly popular) benchmark datasets, partially because good benchmark datasets for CATE estimation are rare. In addition to considering further semi-synthetic (ACIC) datasets with simulated outcomes \cite{hahn2019atlantic, carvalho2019assessing} (or using real outcomes for the untreated and simulating only TEs as in \cite{knaus2021machine}), a promising alternative to hand-crafting (possibly unrealistic) DGPs could be to create benchmarks using a generative approach as proposed in e.g. \cite{neal2020realcause}. Nonetheless, we would expect that the resulting benchmarks would then inherently favor estimators that are most similar to the methods having generated the data. Such `credibility' problems do not arise in standard supervised ML, as they can be be overcome by simply evaluating methods on real data  -- yet, in the treatment effect context, this is usually prohibited by the absence of counterfactuals. The Twins dataset used in \cite{louizos2017causal, yoon2018ganite}, in which twins represent counterfactuals, presents an interesting exception. With this in mind, we consider further \textit{curation and provision of publicly available datasets}, in which (proxies for) counterfactuals can reasonably be inferred from real data, to be a fruitful and crucial opportunity for collaboration of the ML HTE community with data owners and domain experts from more applied fields.

%\begin{itemize}
%\item Redefine thinking of what 'state-of-the-art' actually means; state-of-the-art should be defined within one class of models and one type of setting
%\item Design further careful benchmarks. E.g. realcause focuses on realism but comes at cost of flexibility. 
%\end{itemize}

%\paragraph{For practice} 
%\begin{itemize}
%\item Model selection is difficult due to absence of ground truth. Fit on potential outcomes can give some insight into choice of ML method (forest based versus NN based). 
%\item However, fit on factuals cannot necessarily say much on amount of heterogeneity in TE; two models can have identical fit on POs but very different fit on CATE
%\item recent model selection criteria for CATE have also only been evaluated on ACIC dataset \cite{alaa2019validating} or  IHDP dataset \cite{saito2020counterfactual} -- do these conclusions hold up?
%\end{itemize}

\newpage
\subsection*{Acknowledgements} We thank anonymous reviewers as well as members of the vanderschaar-lab for many insightful comments and suggestions. AC gratefully acknowledges funding from AstraZeneca. 
\bibliography{example_paper}

\begin{thebibliography}{41}
\providecommand{\natexlab}[1]{#1}
\providecommand{\url}[1]{\texttt{#1}}
\expandafter\ifx\csname urlstyle\endcsname\relax
  \providecommand{\doi}[1]{doi: #1}\else
  \providecommand{\doi}{doi: \begingroup \urlstyle{rm}\Url}\fi

\bibitem[Alaa \& van~der Schaar(2018)Alaa and van~der Schaar]{alaa2018limits}
Alaa, A. and van~der Schaar, M.
\newblock Limits of estimating heterogeneous treatment effects: Guidelines for
  practical algorithm design.
\newblock In \emph{International Conference on Machine Learning}, pp.\
  129--138, 2018.

\bibitem[Alaa \& Van Der~Schaar(2019)Alaa and Van
  Der~Schaar]{alaa2019validating}
Alaa, A. and Van Der~Schaar, M.
\newblock Validating causal inference models via influence functions.
\newblock In \emph{International Conference on Machine Learning}, pp.\
  191--201. PMLR, 2019.

\bibitem[Alaa \& van~der Schaar(2017)Alaa and van~der Schaar]{alaa2017bayesian}
Alaa, A.~M. and van~der Schaar, M.
\newblock Bayesian inference of individualized treatment effects using
  multi-task gaussian processes.
\newblock \emph{Advances in Neural Information Processing Systems},
  30:\penalty0 3424--3432, 2017.

\bibitem[Assaad et~al.(2021)Assaad, Zeng, Tao, Datta, Mehta, Henao, Li, and
  Duke]{assaad2020counterfactual}
Assaad, S., Zeng, S., Tao, C., Datta, S., Mehta, N., Henao, R., Li, F., and
  Duke, L.~C.
\newblock Counterfactual representation learning with balancing weights.
\newblock In \emph{International Conference on Artificial Intelligence and
  Statistics}, pp.\  1972--1980. PMLR, 2021.

\bibitem[Athey \& Wager(2019)Athey and Wager]{athey2019estimating}
Athey, S. and Wager, S.
\newblock Estimating treatment effects with causal forests: An application.
\newblock \emph{arXiv preprint arXiv:1902.07409}, 2019.

\bibitem[Athey et~al.(2019)Athey, Tibshirani, Wager,
  et~al.]{athey2019generalized}
Athey, S., Tibshirani, J., Wager, S., et~al.
\newblock Generalized random forests.
\newblock \emph{The Annals of Statistics}, 47\penalty0 (2):\penalty0
  1148--1178, 2019.

\bibitem[Ballman(2015)]{ballman2015biomarker}
Ballman, K.~V.
\newblock Biomarker: predictive or prognostic?
\newblock \emph{Journal of clinical oncology: official journal of the American
  Society of Clinical Oncology}, 33\penalty0 (33):\penalty0 3968--3971, 2015.

\bibitem[Carvalho et~al.(2019)Carvalho, Feller, Murray, Woody, and
  Yeager]{carvalho2019assessing}
Carvalho, C., Feller, A., Murray, J., Woody, S., and Yeager, D.
\newblock Assessing treatment effect variation in observational studies:
  Results from a data challenge.
\newblock \emph{arXiv preprint arXiv:1907.07592}, 2019.

\bibitem[Curth \& van~der Schaar(2021)Curth and van~der Schaar]{curth2020}
Curth, A. and van~der Schaar, M.
\newblock Nonparametric estimation of heterogeneous treatment effects: From
  theory to learning algorithms.
\newblock In \emph{International Conference on Artificial Intelligence and
  Statistics}, pp.\  1810--1818. PMLR, 2021.

\bibitem[Davis \& Heller(2017)Davis and Heller]{davis2017using}
Davis, J. and Heller, S.~B.
\newblock Using causal forests to predict treatment heterogeneity: An
  application to summer jobs.
\newblock \emph{American Economic Review}, 107\penalty0 (5):\penalty0 546--50,
  2017.

\bibitem[Deng et~al.(2009)Deng, Dong, Socher, Li, Li, and
  Fei-Fei]{deng2009imagenet}
Deng, J., Dong, W., Socher, R., Li, L.-J., Li, K., and Fei-Fei, L.
\newblock Imagenet: A large-scale hierarchical image database.
\newblock In \emph{2009 IEEE conference on computer vision and pattern
  recognition}, pp.\  248--255. Ieee, 2009.

\bibitem[Dorie et~al.(2019)Dorie, Hill, Shalit, Scott, Cervone,
  et~al.]{dorie2019automated}
Dorie, V., Hill, J., Shalit, U., Scott, M., Cervone, D., et~al.
\newblock Automated versus do-it-yourself methods for causal inference: Lessons
  learned from a data analysis competition.
\newblock \emph{Statistical Science}, 34\penalty0 (1):\penalty0 43--68, 2019.

\bibitem[Hahn et~al.(2019)Hahn, Dorie, and Murray]{hahn2019atlantic}
Hahn, P.~R., Dorie, V., and Murray, J.~S.
\newblock Atlantic causal inference conference (acic) data analysis challenge
  2017.
\newblock \emph{arXiv preprint arXiv:1905.09515}, 2019.

\bibitem[Hahn et~al.(2020)Hahn, Murray, Carvalho, et~al.]{hahn2017bayesian}
Hahn, P.~R., Murray, J.~S., Carvalho, C.~M., et~al.
\newblock Bayesian regression tree models for causal inference: Regularization,
  confounding, and heterogeneous effects (with discussion).
\newblock \emph{Bayesian Analysis}, 15\penalty0 (3):\penalty0 965--1056, 2020.

\bibitem[Hassanpour \& Greiner(2019)Hassanpour and
  Greiner]{hassanpour2019counterfactual}
Hassanpour, N. and Greiner, R.
\newblock Counterfactual regression with importance sampling weights.
\newblock In \emph{IJCAI}, pp.\  5880--5887, 2019.

\bibitem[Hassanpour \& Greiner(2020)Hassanpour and
  Greiner]{hassanpour2020learning}
Hassanpour, N. and Greiner, R.
\newblock Learning disentangled representations for counterfactual regression.
\newblock In \emph{International Conference on Learning Representations}, 2020.

\bibitem[Hill(2011)]{hill2011bayesian}
Hill, J.~L.
\newblock Bayesian nonparametric modeling for causal inference.
\newblock \emph{Journal of Computational and Graphical Statistics}, 20\penalty0
  (1):\penalty0 217--240, 2011.

\bibitem[Hill(2016)]{acic16}
Hill, J.~L.
\newblock 2016 atlantic causal inference competition: Is your satt where it's
  at?, 2016.
\newblock URL \url{https://jenniferhill7.wixsite.com/acic-2016/competition}.

\bibitem[Holland(1986)]{holland1986statistics}
Holland, P.~W.
\newblock Statistics and causal inference.
\newblock \emph{Journal of the American statistical Association}, 81\penalty0
  (396):\penalty0 945--960, 1986.

\bibitem[Jesson et~al.(2020)Jesson, Mindermann, Shalit, and
  Gal]{jesson2020identifying}
Jesson, A., Mindermann, S., Shalit, U., and Gal, Y.
\newblock Identifying causal-effect inference failure with uncertainty-aware
  models.
\newblock \emph{Advances in Neural Information Processing Systems}, 33, 2020.

\bibitem[Johansson et~al.(2016)Johansson, Shalit, and
  Sontag]{johansson2016learning}
Johansson, F., Shalit, U., and Sontag, D.
\newblock Learning representations for counterfactual inference.
\newblock In \emph{International conference on machine learning}, pp.\
  3020--3029. PMLR, 2016.

\bibitem[Johansson et~al.(2018)Johansson, Kallus, Shalit, and
  Sontag]{johansson2018learning}
Johansson, F.~D., Kallus, N., Shalit, U., and Sontag, D.
\newblock Learning weighted representations for generalization across designs.
\newblock \emph{arXiv preprint arXiv:1802.08598}, 2018.

\bibitem[Kennedy(2020)]{kennedy2020optimal}
Kennedy, E.~H.
\newblock Optimal doubly robust estimation of heterogeneous causal effects.
\newblock \emph{arXiv preprint arXiv:2004.14497}, 2020.

\bibitem[Knaus et~al.(2021)Knaus, Lechner, and Strittmatter]{knaus2021machine}
Knaus, M.~C., Lechner, M., and Strittmatter, A.
\newblock Machine learning estimation of heterogeneous causal effects:
  Empirical monte carlo evidence.
\newblock \emph{The Econometrics Journal}, 24\penalty0 (1):\penalty0 134--161,
  2021.

\bibitem[K{\"u}nzel et~al.(2019)K{\"u}nzel, Sekhon, Bickel, and
  Yu]{kunzel2019metalearners}
K{\"u}nzel, S.~R., Sekhon, J.~S., Bickel, P.~J., and Yu, B.
\newblock Metalearners for estimating heterogeneous treatment effects using
  machine learning.
\newblock \emph{Proceedings of the national academy of sciences}, 116\penalty0
  (10):\penalty0 4156--4165, 2019.

\bibitem[Lipton \& Steinhardt(2018)Lipton and Steinhardt]{lipton2018troubling}
Lipton, Z.~C. and Steinhardt, J.
\newblock Troubling trends in machine learning scholarship.
\newblock \emph{arXiv preprint arXiv:1807.03341}, 2018.

\bibitem[Louizos et~al.(2017)Louizos, Shalit, Mooij, Sontag, Zemel, and
  Welling]{louizos2017causal}
Louizos, C., Shalit, U., Mooij, J.~M., Sontag, D., Zemel, R., and Welling, M.
\newblock Causal effect inference with deep latent-variable models.
\newblock In \emph{Advances in Neural Information Processing Systems}, pp.\
  6446--6456, 2017.

\bibitem[Lu et~al.(2020)Lu, Tao, Chen, Li, Guo, and Carin]{lu2020reconsidering}
Lu, D., Tao, C., Chen, J., Li, F., Guo, F., and Carin, L.
\newblock Reconsidering generative objectives for counterfactual reasoning.
\newblock \emph{Advances in Neural Information Processing Systems}, 33, 2020.

\bibitem[Neal et~al.(2020)Neal, Huang, and Raghupathi]{neal2020realcause}
Neal, B., Huang, C.-W., and Raghupathi, S.
\newblock Realcause: Realistic causal inference benchmarking.
\newblock \emph{arXiv preprint arXiv:2011.15007}, 2020.

\bibitem[Nie \& Wager(2017)Nie and Wager]{nie2017quasi}
Nie, X. and Wager, S.
\newblock Quasi-oracle estimation of heterogeneous treatment effects.
\newblock \emph{arXiv preprint arXiv:1712.04912}, 2017.

\bibitem[Pearl(1995)]{pearl1995causal}
Pearl, J.
\newblock Causal diagrams for empirical research.
\newblock \emph{Biometrika}, 82\penalty0 (4):\penalty0 669--688, 1995.

\bibitem[Pearl(2009)]{pearl2009causality}
Pearl, J.
\newblock \emph{Causality}.
\newblock Cambridge university press, 2009.

\bibitem[Robinson(1988)]{robinson1988root}
Robinson, P.~M.
\newblock Root-n-consistent semiparametric regression.
\newblock \emph{Econometrica: Journal of the Econometric Society}, pp.\
  931--954, 1988.

\bibitem[Rosenbaum \& Rubin(1983)Rosenbaum and Rubin]{rosenbaum1983central}
Rosenbaum, P.~R. and Rubin, D.~B.
\newblock The central role of the propensity score in observational studies for
  causal effects.
\newblock \emph{Biometrika}, 70\penalty0 (1):\penalty0 41--55, 1983.

\bibitem[Rubin(2005)]{rubin2005causal}
Rubin, D.~B.
\newblock Causal inference using potential outcomes: Design, modeling,
  decisions.
\newblock \emph{Journal of the American Statistical Association}, 100\penalty0
  (469):\penalty0 322--331, 2005.

\bibitem[Shalit et~al.(2017)Shalit, Johansson, and
  Sontag]{shalit2017estimating}
Shalit, U., Johansson, F.~D., and Sontag, D.
\newblock Estimating individual treatment effect: generalization bounds and
  algorithms.
\newblock In \emph{International Conference on Machine Learning}, pp.\
  3076--3085. PMLR, 2017.

\bibitem[Tibshirani et~al.(2020)Tibshirani, Athey, Friedberg, Hadad, Hirshberg,
  Miner, Sverdrup, Wager, and Wright]{grf}
Tibshirani, J., Athey, S., Friedberg, R., Hadad, V., Hirshberg, D., Miner, D.,
  Sverdrup, E., Wager, S., and Wright, M.
\newblock grf: Generalized random forests, 2020.
\newblock URL \url{https://github.com/grf-labs/grf. R}.

\bibitem[Wager \& Athey(2018)Wager and Athey]{wager2018estimation}
Wager, S. and Athey, S.
\newblock Estimation and inference of heterogeneous treatment effects using
  random forests.
\newblock \emph{Journal of the American Statistical Association}, 113\penalty0
  (523):\penalty0 1228--1242, 2018.

\bibitem[Yao et~al.(2018)Yao, Li, Li, Huai, Gao, and
  Zhang]{yao2018representation}
Yao, L., Li, S., Li, Y., Huai, M., Gao, J., and Zhang, A.
\newblock Representation learning for treatment effect estimation from
  observational data.
\newblock \emph{Advances in Neural Information Processing Systems}, 31, 2018.

\bibitem[Yoon et~al.(2018)Yoon, Jordon, and van~der Schaar]{yoon2018ganite}
Yoon, J., Jordon, J., and van~der Schaar, M.
\newblock Ganite: Estimation of individualized treatment effects using
  generative adversarial nets.
\newblock In \emph{International Conference on Learning Representations}, 2018.

\bibitem[Zhang et~al.(2020)Zhang, Bellot, and van~der
  Schaar]{zhang2020learning}
Zhang, Y., Bellot, A., and van~der Schaar, M.
\newblock Learning overlapping representations for the estimation of
  individualized treatment effects.
\newblock \emph{arXiv preprint arXiv:2001.04754}, 2020.

\end{thebibliography}
\bibliographystyle{icml2021}

\appendix
\section{Additional implementation details}
We use all random forests with standard hyperparameters as implemented in \cite{grf}; in particular this entails using a very large number of trees (2000). All neural networks are used with standard hyperparameters and components as implemented in \url{catenets}\footnote{Available at \url{https://github.com/AliciaCurth/CATENets}} for \cite{curth2020}; these are in turn based on the hyperparameters used in \cite{shalit2017estimating} for the IHDP experiments. In particular, all networks use dense layers with exponential linear units (ELU) as nonlinear activation functions, are trained with Adam, minibatches of size 100, and use early stopping based on a 30\% validation split. All estimators have 3 representation layers of 200 units, 2 hypothesis layers with 100 units and a final prediction layer, and a small l2-penalty is applied to all weights. Refer to \cite{curth2020} for further detail. Finally, we use RNet without cross-fitting. 

We retrieved the IHDP-100 data from \url{https://www.fredjo.com/}. We retrieved the ACIC2016 competition data from \url{https://jenniferhill7.wixsite.com/acic-2016/competition}; the transformations for Fig. \ref{fig:acictrans} were performed using the script \url{https://github.com/vdorie/aciccomp/blob/master/2016/R/transformInput.R} from the competition R-package. 

\section{Additional results}
\begin{figure*}[!t]
	\centering
	\subfigure[Original IHDP]{\label{fig:ihdp_main_rf}\includegraphics[width=0.9\columnwidth]{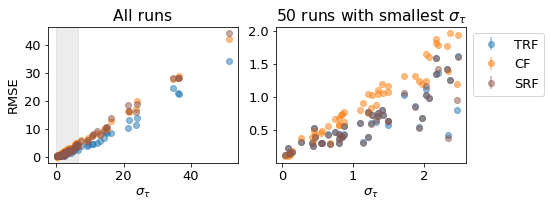}}
    \subfigure[Modified IHDP (Additive TE)]{\label{fig:ihdp_mod_rf}\includegraphics[width=0.9\columnwidth]{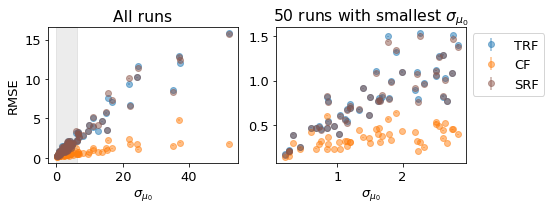}}
    \vspace{-4.5mm}
	\caption{Out-of-sample RMSE of CATE estimation across 100 IHDP draws (original and modified setting)  for forest-based estimators. Averaged across 5 runs, bar indicates one standard error.  Shaded area in left plots indicates area which are zoomed on in right plots.}\label{fig:ihdp_rf}      
\end{figure*}

\begin{figure*}[!t]
	\centering
	\subfigure[Original IHDP]{\label{fig:ihdp_main_net}\includegraphics[width=0.9\columnwidth]{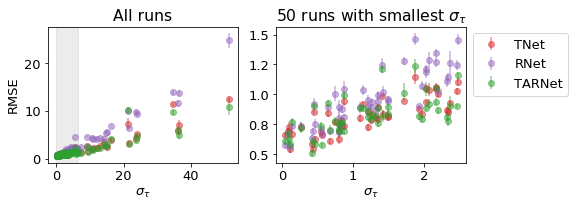}}
    \subfigure[Modified IHDP (Additive TE)]{\label{fig:ihdp_mod_net}\includegraphics[width=0.9\columnwidth]{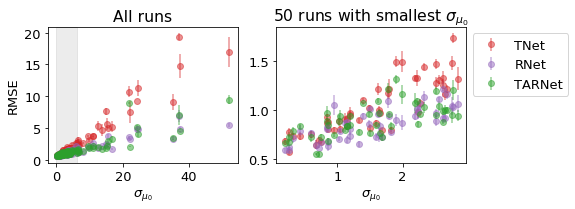}}
    \vspace{-4.5mm}
	\caption{Out-of-sample RMSE of CATE estimation across 100 IHDP draws (original and modified setting)  for NN-based estimators. Averaged across 5 runs, bar indicates one standard error.  Shaded area in left plots indicates area which are zoomed on in right plots.}\label{fig:ihdp_nets}      
\end{figure*}

In Fig. \ref{fig:ihdp_rf} (RFs) and \ref{fig:ihdp_nets} (NNs) we present full results on the two IHDP settings, which we left out of the main text for readability. Similar to the ACIC experiments, performance differences are more striking for RF-based than for NN-based methods.

%To do: count papers at AISTATS, NEURIPS, ICML, ICLR since 2016 which have relied on IHDP datasets, or some of ACIC datasets. 

%ToDO: Make table of empirical evaluation used in related work (ML)

%Contrast with more careful evaluations; e.g. \cite{nie2017quasi}, \cite{kunzel2019metalearners}. 

\end{document}